# A wearable sensor vest for social humanoid robots with GPGPU, IoT, and modular software architecture

Mohsen Jafarzadeh, Stephen Brooks, Shimeng Yu, Balakrishnan Prabhakaran and Yonas Tadesse

## Bibtex

@article{Jafarzadeh2021robots,
title = {A wearable sensor vest for social humanoid robots with GPGPU, IoT, and modular software architecture},
journal = {Robotics and Autonomous Systems},
volume = {139},
pages = {103536},
year = {2021},
issn = {0921-8890},
doi = {https://doi.org/10.1016/j.robot.2020.103536},
url = {https://www.sciencedirect.com/science/article/pii/S0921889019306323},
author = {Mohsen Jafarzadeh and Stephen Brooks and Shimeng Yu and Balakrishnan Prabhakaran and Yonas Tadesse},
}

# A wearable sensor vest for social humanoid robots with GPGPU, IoT, and modular software architecture

Mohsen Jafarzadeh[a,b], Stephen Brooks[a,b], Shimeng Yu[c], Balakrishnan Prabhakaran[c] and Yonas Tadesse[a,b]*

[a] Department of Electrical and Computer Engineering, The University of Texas at Dallas, Richardson, TX 75080 USA

[b] Humanoid, Biorobotics and Smart Systems (HBS) Laboratory, Department of Mechanical Engineering,

[c] Department of Computer Science, The University of Texas at Dallas, Richardson, TX 75080 USA

* Corresponding author. Tel.: +1-972-883-4556; fax: +1-972-883-4659. E-mail address: yonas.tadesse@utdallas.edu

## Abstract

Currently, most social robots interact with their surroundings or humans through sensors that are integral parts of the robots, which limits the usability of the sensors, human-robot interaction, and interchangeability. A wearable sensor garment that fits many robots is needed in many applications. This article presents an affordable wearable sensor vest, and an open-source software architecture with the Internet of Things (IoT) for social humanoid robots. The vest consists of touch, temperature, gesture, distance, vision sensors, and a wireless communication module. The IoT feature allows the robot to interact with humans locally and over the Internet. The designed architecture works for any social robot that has a general purpose graphics processing unit (GPGPU), $I^2C$/SPI buses, Internet connection, and the Robotics Operating System (ROS). The modular design of this architecture enables developers to easily add/remove/update complex behaviors. The proposed software architecture provides IoT technology, GPGPU nodes, $I^2C$ and SPI bus mangers, audio-visual interaction nodes (speech to text, text to speech, and image understanding), and isolation between behavior nodes and other nodes. The proposed IoT solution consists of related nodes in the robot, a RESTful web service, and user interfaces. We used the HTTP protocol as a means of two-way communication with the social robot over the Internet. Developers can easily edit or add nodes in C, C++, and Python programming languages. Our architecture can be used for designing more sophisticated behaviors for social humanoid robots.

## 1. Introduction

Interaction between humans and humanoid robots as well as interaction between many humanoid robots among each other is an open problem that has been studied for many years [1-5]. The response of the robots should be real-time such that people who interact with them [6] do not notice latency. Social humanoid robots can interact via different ways such as touch [7], speech [8], vision [9], etc. In addition to the local interactions, social humanoid robots can interact remotely. By adding Internet of Things (IoT) technology to the robots [10, 11], they can interact with anyone who has access to the Internet (and has permission). Moreover, IoT allows social robots to request information from the Internet and return it to users. Therefore, a software architecture should address real-time interaction through all physical mediums and the Internet.

Developing software architectures for social humanoid robots is challenging because the software architectures consist of many software nodes which require communication with each other. Software architectures are usually limited by hardware. The first generation of social robots only have one or few microcontrollers. For example, Robota used an 80C51 microcontroller family [12]. As a result, such a robot cannot use image data because the microcontroller does not have enough memory. Also, the speed of the microcontroller does not allow developers to write a comprehensive and low latency software/codes. If the codes do not run within a reasonable time, the consumers experience awkward moments. The second generation of social humanoid robots have a general-purpose and more powerful CPU on the mainboard and run an operating system. For example, Zeno has a Vortex86DX x86 CPU and runs Debian Linux [13]. This generation of social humanoid robots enables developers to write complex programs. However, they have limited power of parallel computing.

Graphics processing units (GPUs) were originally attached to video cards designed specifically to accelerate graphics rendering and display [14]. GPU architectures employ two important execution paradigms, high floating-point (FP) execution bandwidth and hardware multithreading [15]. Recently, the traditional fixed-function graphics pipeline has evolved into a much more flexible and unified many-core architecture, and non-graphics data-parallel languages have been introduced to program these cores [16]. The GPUs that can be used for non-graphics applications with non-graphics programming languages or libraries are called general-purpose graphics processing units (GPGPUs) [17]. The term “general-purpose” refers to “non-graphics domains”. In other words, GPGPUs can be used for both graphical processing and scientific computing. The compute unified device architecture (CUDA) is the scheme by which NVIDIA has built GPUs that can perform both traditional graphics-rendering tasks and general-purpose tasks [18]. The early versions of the CUDA do not have all the necessary features to be used as a comprehensive GPGPU in robotics. In this paper, we only consider the CUDA with compute capability 6.0 and higher as GPGPU. NVIDIA Jetson TX2 (released on May 2017) [19] and NVIDIA AGX Xavier (released on September 2018) [20] are examples of CUDA with compute capability 6.2 and 7.1 respectively. By this definition of the GPGPU, most of the social humanoid robots in the literature did not have any embedded GPGPU. Few researchers used external GPGPUs for their social humanoid robots. This is because the embedded GPGPU was released recently (May 2017), and there was not any GPGPU at the time that they designed their robots. Here, for the first time, we used an embedded GPGPU for a social humanoid robot.

The GPGPUs play an important role in training and inference of deep neural networks. Researchers widely use deep neural networks [21] for perception, decision, and control. For examples, the following have been presented imitation learning [22], image understanding (object recognition [23], object detection [24], object tracking [25], and object segmentation [26]), speech recognition [27], natural language processing [28], end-to-end visuomotor learning [29], controlling prosthetic hands with speech [30] and EMG [31], etc. In addition, GPGPU can boost speeds of recurrent neural networks [32, 33] and recently recurrent neural networks (RNNs) are shown relevance in robotics for solving time-varying problems [34-36]. After the success of deep reinforcement learning in achieving human performance in playing Atari [37], deep reinforcement learning is used in the humanoid robotics field. The next generation of social humanoid robots should use an architecture that enables robots to use deep reinforcement learning. In addition, the latest generation shows complex behaviors such as managing trust [38], learning from human-human interaction [39], inferring social situations [40], interaction based on the context [41], and learning from human [42].

IoT is a non-robotic technology that improves robots. IoT is a concept reflecting a connected set of anyone, anything, anytime, anyplace, any service, and any network [43]. IoT can be thought of as the interconnection of uniquely identifiable smart objects and devices within today's Internet infrastructure with extended benefits. These benefits typically include the advanced connectivity of these devices, systems, and services that go beyond machine-to-machine (M2M) scenarios [44]. Sensors and data analytics technologies from IoT give robots a wider situational awareness that leads to better task execution [45]. Processing the data streams generated by billions of IoT devices in a handful of centralized data brings concerns on response time latency, massive ingress bandwidth needs and data privacy [46]. Therefore, working in IoT does not explicitly address the requirement of the Internet of Robotic Things (IoRT) to exchange continuous streams of data while interacting with the physical world [46].

Cloud computing is another non-robotic technology that has impacts on robotics. The cloud paradigm was adopted by the robotics community, called cloud robotics [47, 48]. Cloud robotics is a system that relies on the "Cloud Computing” infrastructure to access a vast amount of processing power and data to support its operation [49]. Robots can use a cloud for offloading resource-intensive tasks for sharing of data and knowledge between robots and for reconfiguration of robots following an app-store model [46]. Although there is an overlap between cloud robotics and the Internet of Robotic Things (IoRT), the former paradigm is more oriented towards providing network-accessible infrastructure for computational power and storage of data and knowledge, while the latter is more focused on M2M communication and intelligent data processing [46].

Many organizations and institutes have designed expensive humanoid robots for specific tasks. NASA designed Robonaut 2 [50], other institutes presented HRP-4 [51], Romeo [52], Kenshiro [53], Surena III [54], Atlas [55], Talos [56], and DRC-Hubo [57]. The detailed information of these advanced robots is shown in supplementary file in Fig. S1. All the aforementioned robots are advanced and highly capable in terms of hardware performances and functions. However, they are very expensive and most of their sensing units, electrical connections and processing boards are integral parts of the robot, which inhibit modularity of the sensing unit and interchangeability. Therefore, a wearable sensor system that can be integrated into a stretchable garment that fits one robot and can be easily used in another robot is needed, which ultimately advances the field. Wearable sensor vests are common in the healthcare area and for many years, researchers have worked on devices that measure human characteristics such as ECG, EMG, sweat rate, etc. (Supplementary files Fig. S2 and S3). In this regard, there are many approaches the robotic community can adapt. Wu et al. designed as a wearable sensor node with solar energy harvesting [58]. In the energy sector, Safety++ is an IoT ecosystem of connected, wearable elements aimed at improving safety in the workplace [59]. Wearable garments for infants have been studied and designed by many researchers [60]. Jutila et al. designed a vest to track seven to nine-year-old students [61]. BIOTEX is a European Union funded project that aims at developing textile sensors to measure physiological parameters and the chemical composition of body fluids with a particular interest in sweat [62]. Wearable 2.0 is a washable garment [63]. Nokia designed CHASE LifeTech FR for firefighters [64]. Carre Technologies Inc. (Hexoskin) [65], Enflux Inc. [66], AiQ (Motion) [67], and Xenoma (E-Skin) [68] made garments for bodybuilders. Ohmatex designed Safe@Sea for fishman [69]. More discussion on this topic is given in the supplementary file and summarized in Fig. S2. All the wearable systems discussed above lend a good lesson for social robots to be interactive and interchangeable. However, we cannot directly transfer these applications to social robots, we need to modify the hardware and the software architecture.

In the design of the proposed vest (Fig. 1- 4), we have tried to reduce the cost as much as possible. Therefore, we made a tradeoff between cost and performance. Our proposed vest has a hardware and software that works together to provide a unique solution to social humanoid robots. The software consists of modular systems (to be discussed later) and the hardware includes sensors, controllers and an embedded GPGPU kit. The specification of our prototype is shown in Table 1, the first column that is a low cost (~$1000 material cost). The current technology does not satisfy all of our required parts and performance as well as cost limitations. Therefore, researchers can use the current design and further improve the vest in the future. An improved vest can be designed by using hardware according to Table 1, column 2. These are based on the available hardware in the market. It is estimated that this may cost more than 15 times higher than the low-cost one. In the following, we describe the ideal vest that should be designed. The hardware may not be available in the market for an ideal vest. Designers may have their own opinion about what an ideal vest entails. In other words, the features of an ideal vest are debatable. The ideal vest software should have the following:

- Monitoring and control over Internet
- Search and use information on the Internet
- GUIs for users and developers
- Color 3D point cloud object detection (processed locally, not IoT) from fusion of cameras and Lidar,
- Text to speech (natural accent) (local, not IoT),
- Accurate speech to text (local, not IoT),
- Attention activation, instead of timing base architecture that takes continuous power

- Reinforcement learning for behavior ( for lifelong learning)
- Reconfigurable architecture by learning while maintaining ethical and safety standards
- Adaptive behavior with respect to application (child interaction, adult interaction),
- Personalizablity according to psychologist instruction

Table 1. Hardware comparison of our prototype and a better prototype with current technology.

| Hardware | Our prototype (low cost) | Better option / existing in the market |
|---|---|---|
| **GPGPU Kit** | NVIDIA Jetson TX2 ($5 \times 10^{11}$ Flops, 64-bit FPU, 170 x 170 x 51 mm) | NVIDIA AGX Xavier ($10^{12}$ FLOPS, 64-bit FPU, 105 x 105 x 85 mm) |
| **Distributed controller** | N/A | Pyboard D (STM32F767) and 32-bit ADC (Analog Devices Inc. AD7177-2BRUZ-RL7), 32-bit DAC (Cirrus Logic Inc. CS43131-CNZR) |
| **Memory** | 32 GB | 4 TB |
| **Internet Communication** | Wi-Fi | 5G |
| **Microphone** | 8-bit unidirectional | 24-bit cardioid (Audio-Technica AT2020USBi) |
| **Camera** | Webcam Logitech C922x Pro USB 2.0 | 2 x FLIR Chameleon3 CM3-U3-13Y3C-CS |
| **Range sensor** | Ultrasonic SFR02 and 1D ToF | Puck VLP-16 Lidar |
| **Odometer** | N/A | IMU (Bosch BNO055), AHRS (Yost Labs 3-Space), and GPS (Adafruit Ultimate). |
| **Tactile Sensors** | Interlink FSR 406 and capacitive touch NXP MPR121 | FSR FlexiForce A502 and capacitive touch NXP MPR121 |
| **Robust to** | - | Shock |
| **Hardware visibility to human** | Visible | Camouflage |
| **Wiring** | External copper wire | Internal copper wire |
| **Style** | Deep Pressure Vest | Shirt |

There are many research gaps in developing social humanoid robots. Acceptability, fault tolerance, decision making, and perception are four important, challenging research gaps. One of the most challenging research gaps is how to design a social humanoid robot that is both culturally and socially acceptable and people have a willingness to engage with it. The second research gap is fault tolerance, which allows the social humanoid robots to keep working even when components fail and the degree to which robots can continue their missions when failures or other unforeseen circumstances occur. In IoT, redundancy is key to fault tolerance. IoT enables redundancy of sensors, information, and actuation [46]. However, too much redundancy may not be possible in social humanoid robots. The third research gap is continuous and real-time learning and decision making. Recent developments in GPGPU and TPU along with new AI methods such as deep reinforcement learning improves the ability of social humanoid robots. However, social humanoid robots still cannot make decisions in complex environments such as collaborating with humans without changing the code or learning a new human language (English, Farsi, Arabic, etc.). Perception is the fourth challenging research gap. Sensor fusion, object detection, and recognition, processing time, and using Internet information are examples of problems in perception.

Designing a modular open-source software architecture (four-clause BSD license) that works with an affordable, customizable wearable, GPGPU and IoT based sensor vest for social humanoid robots is the main motivation as well as the contribution of this paper. Our goal is providing a general framework to be used as a basis for developing low cost, socially interactive humanoid robots (Fig.1). The significant features of this software architecture along with the wearable GPGPU and IoT-based vest enable social robots to interact with humans and other robots both locally and remotely via the Internet. The proposed architecture for the wearable IoT vest provides the following:

1) A framework that isolates the behavior nodes from other nodes (service node, hardware node, bus manager node), such that users can design complex behavior without knowing/concerning about the whole system (Fig. 1).
2) A solution to add general-purpose graphics processing unit (GPGPU) and IoT capability to social humanoid robots. Administrators can monitor and control the social robots over the Internet. This solution consists of related nodes in the robots, a RESTful web service, and a user interface (Fig.2).
3) A provision for audio-visual interaction. It consists of three subsystems: speech to text, text to speech, and image understanding. These provisions require an embedded GPGPU. Each subsystem can be modified/optimized independently.
4) A constitution for adding/managing sensors and actuators to the social robots. This constitution rules the bus manager such as $I^2C$ and SPI (section 3).
5) A hardware architecture for accomplishing the aforementioned features (Fig. 3). The proposed hardware consists: a wearable vest, a GPGPU kit with $I^2C$, SPI, and USB (3.0 or higher) busses that can run Linux, a wireless Internet communication either Wi-Fi or 5G, USB cameras, a microphone, a speaker, and sensors( touch, force, temperature, and distance) that communicate via $I^2C$ or SPI.

This paper is organized as follows. Section 2 briefly describes the related works. The proposed software architecture is explained in section 3. Then, a case study is described in section 4 including hardware details, software nodes, and experimental results of the low-cost prototype. Discussion on the proposed architecture is stated in section 5. The final section is a conclusion for the proposed architecture for social humanoid robots. A supplementary file is also provided that explains in greater detail in some areas with more detailed specifications, figures and explanations.

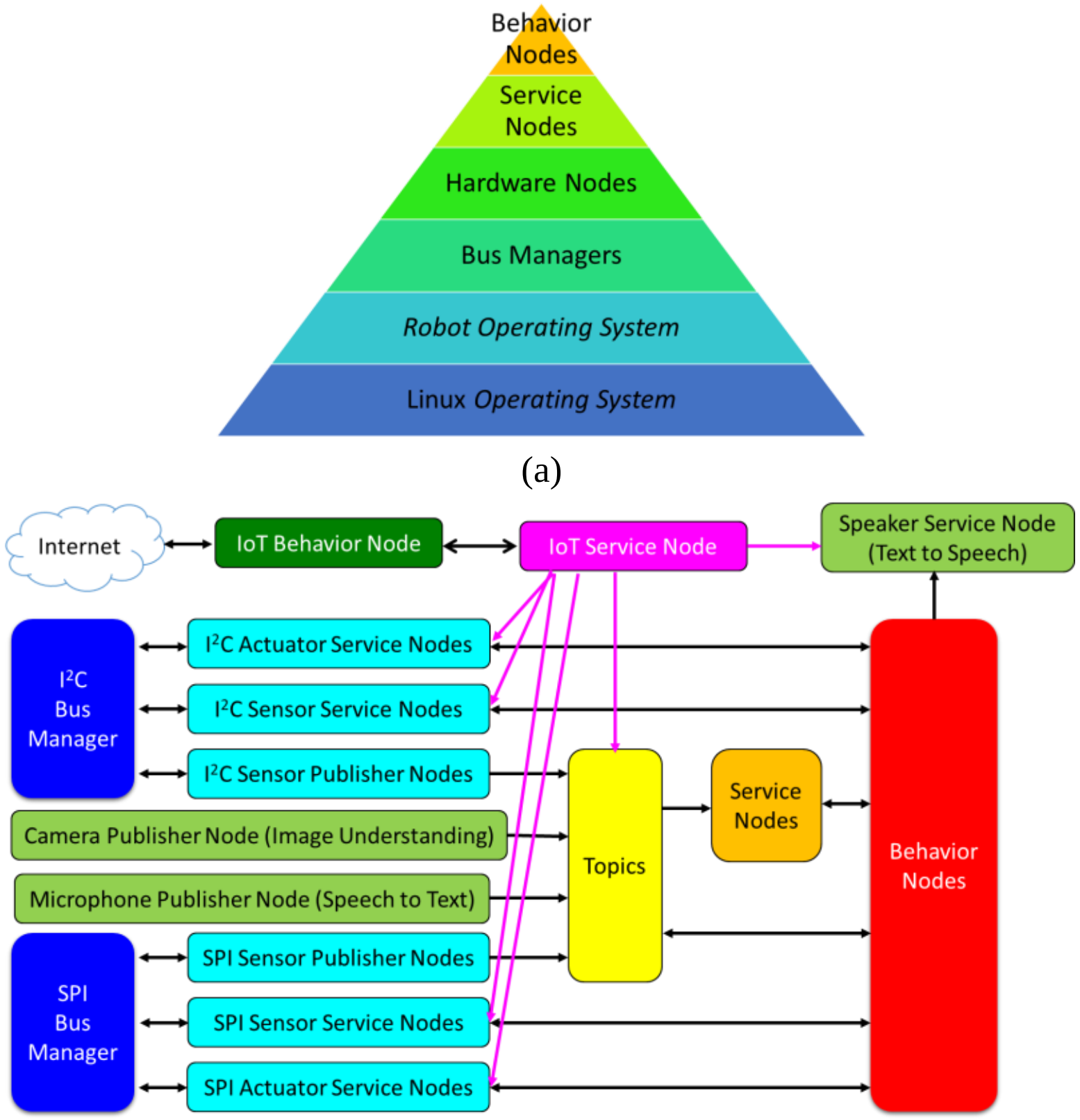

(a)

(b)

**Fig.1**. The proposed software architecture: (a) level pyramid (b) nodes arrangement.

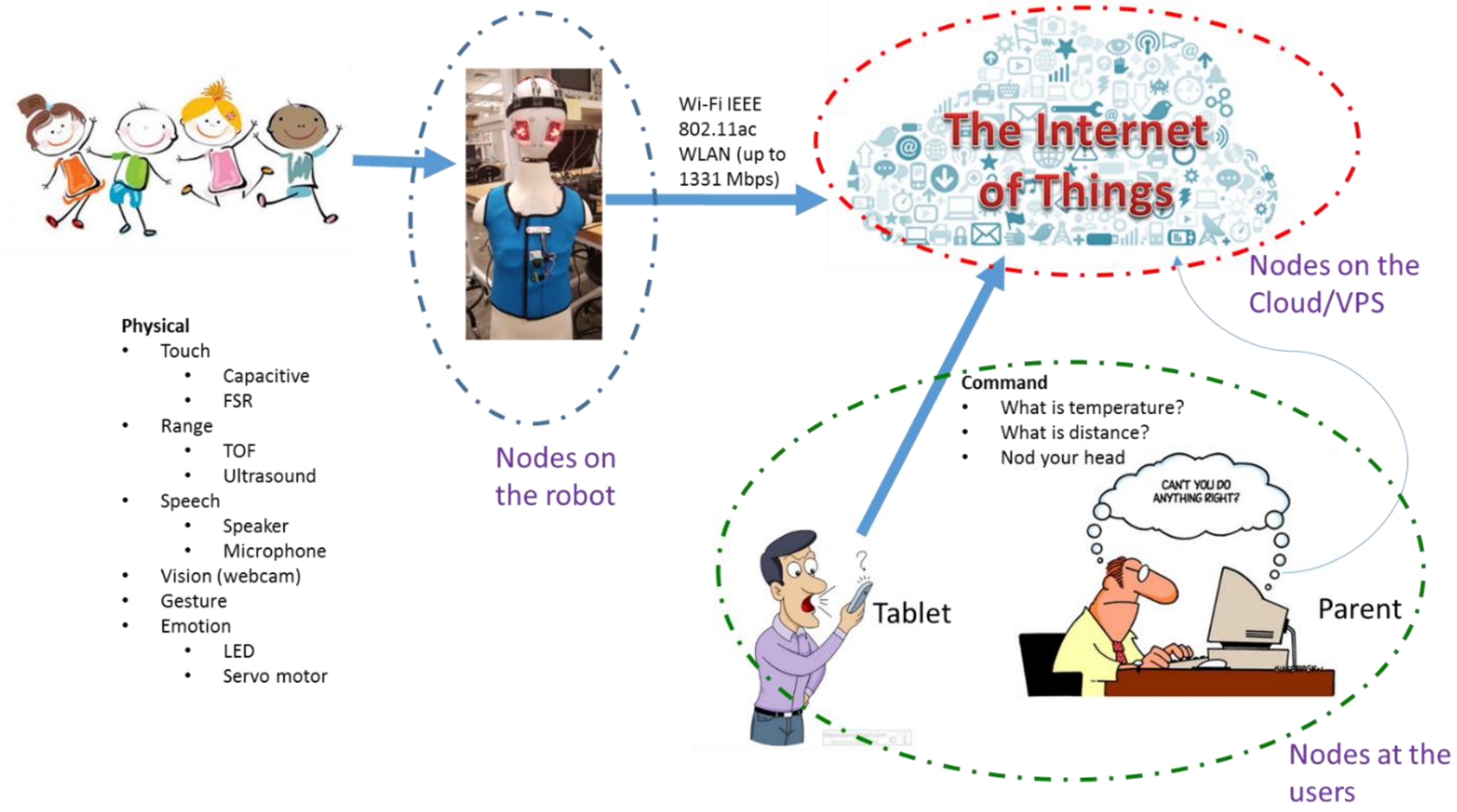

**Fig.2.** An illustration of how a vest for a social robot work in a typical Internet of Things (IoT) framework.

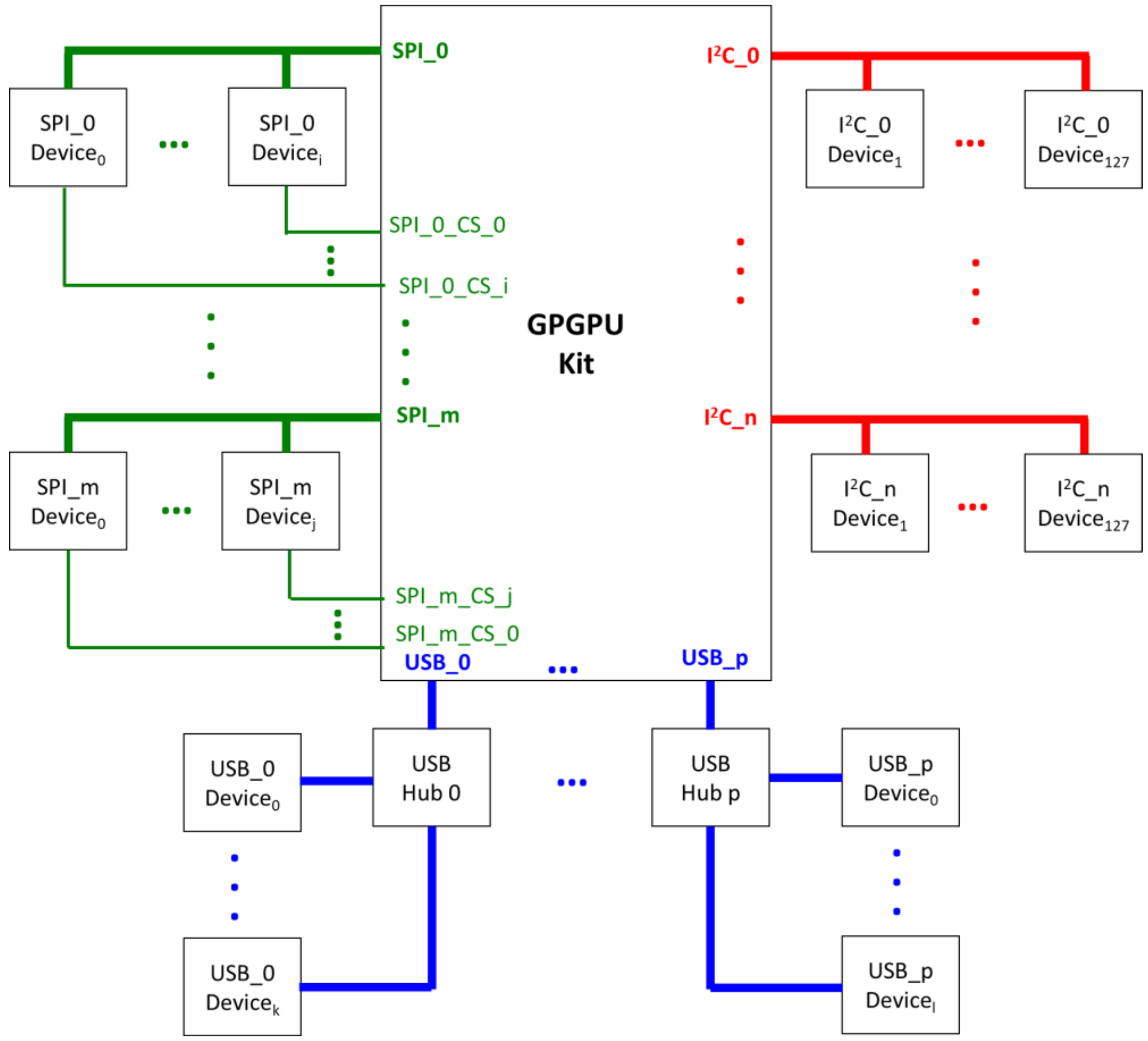

Fig. 3. The hardware architecture that works with the proposed software architecture.

## 2. Related Works on Small Social Robots

Researchers have developed many humanoid robots that are designed to be used as social interactive robots. Many humanoid robots are designed for other goals. In the following, some of the popular social robots are introduced. Robota is a series of social robots. The robots are doll-like and designed to encourage children to play and interact with them. The first version of the Robota was developed in 1997 [12] and uses the 80C51 microcontroller family. The last version of Robota has a Pocket PC (Compaq iPAQ-3850 Intel Strong ARM SA-1110 32-bit RISC Processor 206 MHz, with 64 Mb of RAM) [70]. A motor controller board (Microchip PIC16F870) and a sensor board (Microchip PIC16F84) connected by a serial port to the Pocket PC. The Pocket PC has a 160x120 pixel camera with a rate of 15 FPS. The camera of Robota enables tracking of the hands and nose of a user.

KASPAR is another social humanoid robot series that is popular in the study of children diagnosed with autism. KASPAR stands for Kinesics and Synchronization in Personal Assistant Robotics. The first version was released in 2005 [71, 72]. The KASPAR can be used in two modes: remotely controlled as well as autonomous operation. In a variety of projects, KASPAR operates autonomously. An application programming interface (API) provides access for programmers to develop custom programs and access to open-source robot software produced under the "yet another robot programme". In applications involving children with autism, a remote control was used to operate KASPAR. It is made of a standard wireless keypad with 20 keys. Different keys were programmed to activate different behaviors in KASPAR, e.g. left/right arm drumming, waving, different postures, etc. Operators can easily execute and develop programs for the robot using a graphical user interface (GUI), which runs on any Microsoft Windows or Linux computer. This interface has been used in human-robot interaction scenarios when an experimenter remotely controlled the robot from a laptop. [73]. The sense-think-act architecture is used to enable KASPAR to display goal-oriented adaptive behaviors while playing, communicating, and collaborating with children [74].

Zeno is an emotional robot with expressive skin developed by Hanson Robotics company [13, 75, 76]. The RoboWorkshop software allows editing and creation of Zeno .xml animation files containing servomotor position information. Automatic lip-syncing text-to-speech software is also provided. However, there is no software, similar to NAO or iCub, available to program Zeno to perform complex behaviors [13]. In addition, Hanson Robotics has made RoboKind API available with extensive Java libraries to communicate with Zeno [13].

ACTROID-F is a social humanoid robot which is similar to a real human and can exhibit various facial expression [77]. The android is connected to a notebook PC via USB cable. A webcam, which captures a facial image of a target person for face recognition, is also connected to a notebook PC. A microphone is connected to the notebook PC via a voice changer. The android's voice that is generated from speech synthesizer or the voice changer is output from a speaker behind the android. All of the degrees of freedom can be manually controlled by sliders in the graphical interface. Unconscious human motions such as blinking, gaze movements, head movements, and breathing were implemented on the android. The eyes show one of the pre-defined motions such as single blinking, double blinking, and gaze movement (vertical or horizontal) stochastically with intervals of 2 to 3 seconds. The graphical interface for the operator has push-buttons corresponding to pre-determined motions. These motion events from the operator are executed in priority to the synchronized motion.

NAO robots are commercial humanoid robots, which are often used in standard platform league (SPL) of Robocup competition [78]. Some versions of NAO robots are designed for use in education and research [79]. NAO robots have been used in social experiments such as teaching [80], autism [81], hotel's information [82], etc. NAO robots are equipped with NAOqi 2.0 which is a branch of Linux Gentoo. Nao can be programmed in two ways, using GUI and using APIs that are provided through a software developer kit. The APIs enable the developers to write code in C++, Python, Urbi and .Net.

The iCub is an open-source humanoid robotic platform designed explicitly to support research in embodied cognition [83-85]. iCub has been used as a social robot in [86, 87]. The robot contains motor amplifiers, a set of DSP controllers, a PC104-based PC, and analog to digital conversion cards. The CAN bus is employed to communicate with the PC104 CPU. Time-consuming computation is typically carried out externally on a cluster of machines. The communication with the robot occurs via a Gb Ethernet connection. The iCub software was developed on top of Yet Another Robot Platform (YARP) [88, 89]. YARP is a set of libraries that support modularity by abstracting two common difficulties in modularity in algorithms and in interfacing with the hardware.

Buddy is a low cost, 3D printed humanoid robot that is controlled by an Arduino Uno board [90, 91]. The board has an Atmel ATmega328P microcontroller (32 kB flash memory, 2 kB SRAM, and 1 kB EEPROM). The computational ability of the robot is very low and just can do some basic tasks such as dancing or playing music, as the focus of the robot was on the 3D printed structure and facial expression. The robot has a Wi-Fi IP camera that captures images and sends the images to a cloud. A PC can be connected to the cloud and use the camera images to process them and extract information. In addition, the PC can send commands to the robot through an IoT board (Particle Photon).

HBS1 is another 3D printed humanoid robot that is controlled by a USB2Dynamixel and a Pololu mini Maestro servo controller [92, 93]. The robot has two cameras with IEEE 1394 in the eyes. The main program has been written for the desktop PC in C++. HBS 1.2 is the next version of HBS robots that has only the upper body [94]. The HBS 1.2 controller is a combination of a desktop computer (Intel Core i7) and a microcontroller board (Nucleo STM32F767zi). The desktop program is written in MATLAB and it controls the Dynamixel servo motors via USB2Dynamixel adapter and RC servo motors via Pololu mini Maestro servo controller. The desktop program also reads and displays the two IEEE 1394 bus in the eye and ZED stereo camera in the chest. The Nucleo STM32F767zi is programmed in C using GNU Arm Embedded toolchains (GCC). The Nucleo board controls artificial muscles in its hands. The communication between the desktop computer and microcontroller board is Universal Asynchronous Receiver Transmitter (UART) [94]. The third version, HBS 2.0 is the newer generation of HBS robot which replaces the desktop computer and the microcontroller board with an embedded GPGPU (NVIDIA Jetson TX2 developer kit) [95]. It has a ZED stereo camera in the chest. The NVIDIA Jetson TX2 developer kit controls Dynamixel by USB2Dynamixel, RC servo by Pololu mini Maestro servo controller, and the artificial muscles by a combination of PWM generator (Adafruit PCA9685) and MOSFET boards (IRF3708). The HBS 2.0 can be programmed in C, C++, and Python.

# 3. Proposed Software Architecture

With all distributed systems, there must be a way to connect subsystems together in order for each of the subsystems to communicate or interact with each other. A communication system is often one of the first needs to arise when implementing a new application in robotics. Here, the Robotic Operating System (ROS) [96] is chosen as the foundation for the messaging system that provides the interconnects between the software sub-modules. For instance, when a person interacts with the humanoid by hugging its waist, the software module that interprets the voltage across our pressure sensors in the area must be able to send that data to another sub-module that knows what the appropriate response should be, which may be to speak to the person or hug back. The proposed software architecture is comprised of many nodes. Therefore, code complexity is reduced in comparison to monolithic systems. A node is an executable file that uses ROS to communicate with other nodes. ROS is an operating system on another operating system (Linux Ubuntu). Developers can write nodes either in C++ or Python. ROS starts with the ROS Master. The Master allows all other ROS pieces of software (nodes) to find and talk to each other. ROS was designed to be as distributed and modular as possible. The ROS middleware provides publish/subscribe anonymous message passing, recording and playback of messages, request/response remote procedure calls and distributed parameter system. Nodes are combined together into a graph and communicate with one another using streaming topics, Remote Procedure Call (RPC) services, and the Parameter Server. Implementation details are also well hidden as the nodes expose a minimal API to the rest of the graph and alternate implementations, even in other programming languages, can easily be substituted. Messages are the ROS data type used when subscribing or publishing to a topic. Nodes can publish messages to a topic as well as subscribe to a topic to receive messages. There can be multiple publishers and subscribers to a topic. Topics are intended for unidirectional, streaming communication. In general, nodes are not aware of whom they are communicating with. Instead, nodes that are interested in data subscribe to the relevant topic while nodes that generate data publish to the relevant topic. Nodes can also provide or use a Service. The asynchronous nature of publish/subscribe messaging works for many communications need in robotics, but sometimes one may want synchronous request/response interactions between processes. Nodes that need to perform remote procedure calls, e.g. receive a response to a request, should use services instead. The ROS middleware also provides a way for tasks to share configuration information through a global key-value store. This system allows developers to easily modify their task settings and even allows tasks to change the configuration of other tasks. To write a new node, the developer does not need to be familiar with the whole system. The modularity of ROS allows developers to add nodes or entire packages easily with minimal effect on the system. A group of developers can work parallel, each writing their own modules without having to be blocked by another module's ongoing development. Each node can be enabled or disabled independently, therefore, adding a new sensor is easy. If one node crashes, other nodes that are not dependent on it can continue working.

In the proposed architecture (Fig. 1), the nodes are categorized into four types: bus manager nodes, hardware nodes, service nodes, and behavior nodes. Hardware nodes typically have one function, which is to provide monitor and control (M&C) of the hardware of the robot such as sensors and motors. Behavior nodes are responsible for interpreting data from sensors to drive interactions. For example, one behavior node is the movement tracking behavior to hold eye-contact, which will continuously read the data published from the proximity sensors. When a condition is met (for example an object is now not directly in front of the humanoid), it will send a request to the servo motor in the neck to turn a specified number of degrees right or left. Service nodes only perform tasks when another node requests that it perform its function. We used service nodes as a function that is accessible from any other node in the ROS architecture. An example of this is the speaker service node. It simply takes a string input from another Node (almost always a behavior node) and it calls upon the text to speech (TTS) application to send the speech to the speaker.

The proposed software architecture provides the Internet of Things (IoT) functionality for social humanoid robots. There are two foundational features for IoT implementation. First, we wanted to be able to read real-time sensor data from the social humanoid robot. Second, we wanted to be able to control the social humanoid robot over the Internet. These two features are the foundation for a virtually endless expansion of M&C of the humanoid. There are four components in our IoRT architecture: a behavior node, a service node, and a public REST API hosted on a public cloud or VPS and a remote client user interface (UI) application. The behavior node continuously sends HTTP GET requests at as fast a rate as possible via the API, which is asking for any updates from the remote client. An update may be any command that the remote client application has sent to the RESTful web service since the last request. If a command is detected in the HTTP response, the IoT behavior node will determine which of the valid commands it has received and immediately sends a request to the IoT service node to delegate the appropriate action to any other service nodes. The API is built upon the Python Flask web micro-framework and can be hosted on any public-facing server. The API is also capable of accepting and differentiating between commands to multiple robots. Each robot can be assigned a unique ID and the behavior nodes will know if the command is meant for its humanoid. This extends also to the remote client application. A single client application could provide M&C for multiple humanoids as well as multiple geographically separated remote clients to control the same humanoid if desired.

The proposed software architecture provide a constitution for adding/managing sensors and actuators. The inter-integrated circuit ($I^2C$) and the serial peripheral interface (SPI) protocols are at the low-end of the communication protocols [97]. Both SPI and $I^2C$ offer good support for communication with low-speed devices, but SPI is better suited to applications in which devices transfer data streams, whereas $I^2C$ is better at bus topology (routing and resources). A major advantage of the $I^2C$ bus is the efficient use of an important and limited embedded system resource - I/O ports [98], it allows a number of sensors to be connected in a bus. Each hardware node is able to specify which bus that it wishes to create a request for and the $I^2C$ manager node constructs a job object entirely based on the requesting node's inputs. All access to each $I^2C$ and SPI buses is brokered by a single software module named the "Bus Manager." These modules maintain two queues, one for pending requests and one for completed requests. When a read or write request is made by any hardware nodes with access to the specific bus manager, the requesting node will continually ask the bus manager if their job request has been completed. Once that Bus Manager shows that the specified job has been completed, it will return the requested data (if any) to the requesting node that might be waiting for a response and the process will be repeated.

Image understanding is an important subsystem of social humanoid robots. Face detection and recognition, object detection and tracking, gesture and emotion recognition, image segmentation, and 3D reconstruction (mapping) are examples of image understanding subsystems. In the proposed architecture, we used the embedded GPGPU of the robot to run the image understanding module in real-time. The results of image understanding subsystem are published in related topic to use in high-level behavior.

Speech is one of the most basic forms of communication that connects us as humans; therefore, we wanted our humanoid to be able to respond to speech from any interactor. The proposed architecture provides a speaker service node (speech-to-text) and a microphone publisher node (text-to-speech nodes). Advanced Linux Sound Architecture (ALSA) provides audio functionality to the Linux operating system and kernel-driven sound card drivers. ALSA also bundles a user space driven library for application developers. In the proposed architecture, we used ALSA high-level API development in the microphone publisher node to enable direct (kernel) interaction with sound devices through ALSA library.

The advantage of the proposed software architecture is that it provides (1) IoT technology, (2) GPGPU nodes (to use deep learning methods), (3) I2C and SPI bus managers, (4) speech interaction (speech to text and text to speech) nodes, (5) image understanding, and (6) isolation between behavior nodes and other nodes. The proposed software architecture handles all the six aspects, which is not available at the same time in other social humanoid robot software architecture. For example, YARP and NAO software architecture do not provide IoT node, $I^2C$ and SPI bus managers, and GPGPU nodes.

# 4. Case Study

## 4.1. Hardware of our low cost prototype

The proposed software architecture can be run on any social humanoid robot that has (1) embedded GPGPU, (2) Linux operating system, (3) robot operating system (ROS), (4) Internet communication hardware, (5) $I^2C$/SPI bus(es). In this section, we explain an example. In this case study, we used a NVIDIA Jetson TX2 developer kit [19], which supports all of the five conditions. The board is compared with others in the supplementary file (Table S1). Fig. 4 shows the hardware architecture and interconnection. Sensors and actuators are selected and the details are provided in a supplementary file (Table S2, Fig. S4). We used a USB sound adapter to connect the microphone and speaker to the robot. To control the servo motor, we used a Pololu mini Maestro twelve channels servo controller in USB communication (Virtual COM port) mode. We designed a two-layer PCB for RGB LEDs on the face of the robot. Each PCB has 3 RGB LEDs (Kingbright WP154A43VBDZGWCA) and 3 MOSFETs (Infineon IRLB8314PBF), i.e. each MOSFET connects to pin 1 of the 3 LEDs. To control the LED, we used a NXP PCA9685 PWM generator (16 channels, with 12-bit resolution, and programmable frequency 24 Hz to 1526 Hz). We used two types of range sensors that communicate via $I^2C$. For long range, we used an ultrasonic range sensor (Devantech SRF02) and for short range, we used three Time of Flight (ToF) distance sensors (STMicroelectronics VL53L0X). The SRF02 has a minimum measurement distance of 180 mm. STMicroelectronics VL53L0X measures 30 mm to 1.2 meters in default mode. We used a gesture sensor from XYZ Interactive Technologies, which communicates via $I^2C$. It is able to detect hand movements such as swipe, air push, hover, circle, and wave. We used a capacitive touch-sensing unit (NXP MPR121), which can handle up to 12 individual touchpads. It is connected to copper sheets to expand the sensing area. For sensing force, we used Interlink FSR 406. FSR stands for Force Sensing Resistor. It exhibits a decrease in resistance with an increase in applied force to the surface of the sensor, the force can be as low as 0.25 N and as high as 10 N. We used a 16-bit Delta-Sigma (ΔΣ) analog to digital convertor (TI ADS1115) to read the FSRs. We used a Microchip Technology MCP9808 temperature sensor (with a typical accuracy of ±0.5°C) for sensing the surrounding environment.

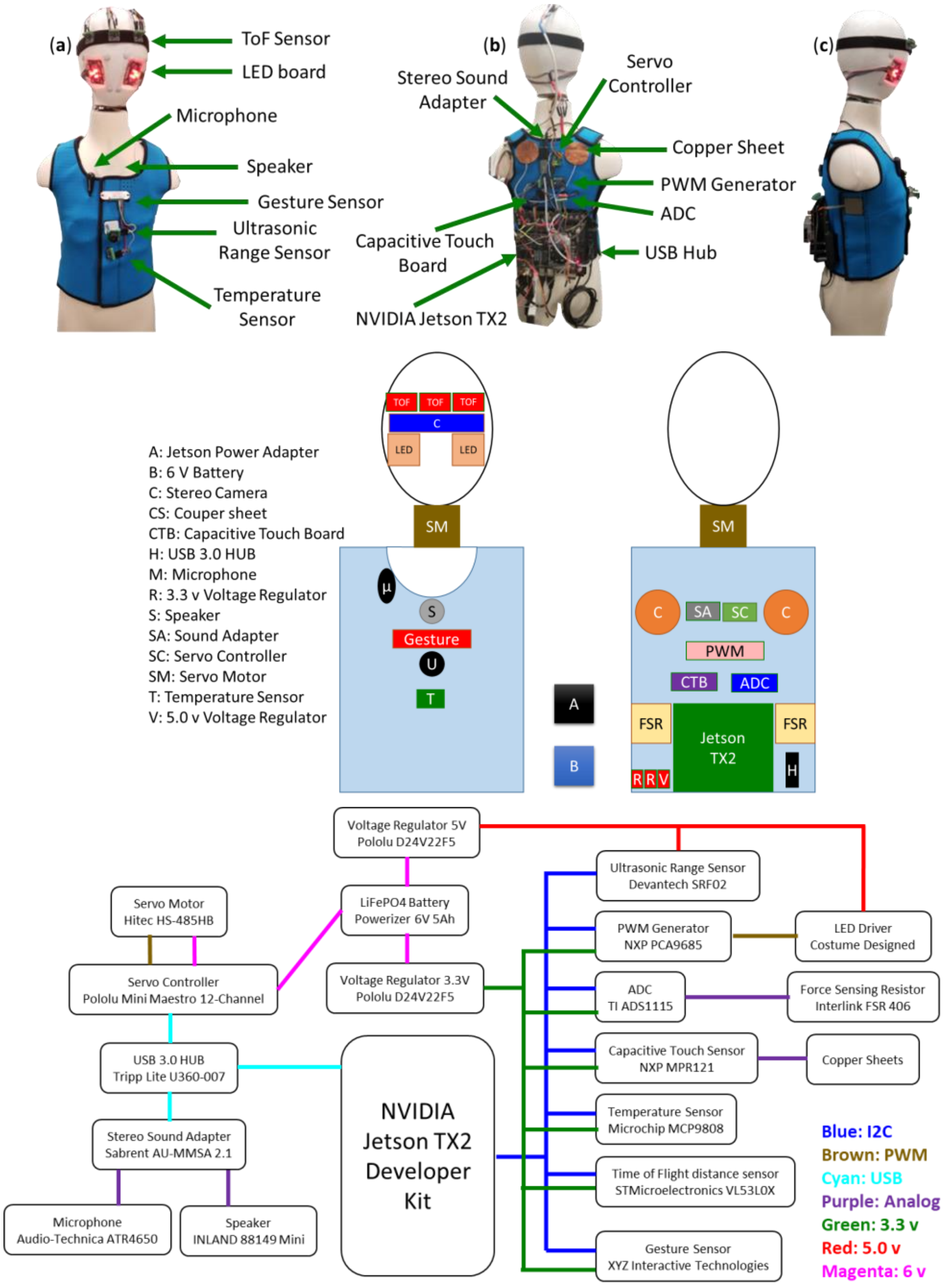

**Fig. 4.** Hardware of the our prototype vest which is mounted on a mannequin as a case study: (a) photograph of the prototype front view, (b) back view, (c) side view, (d) physical diagram, and (e) connection diagrams.

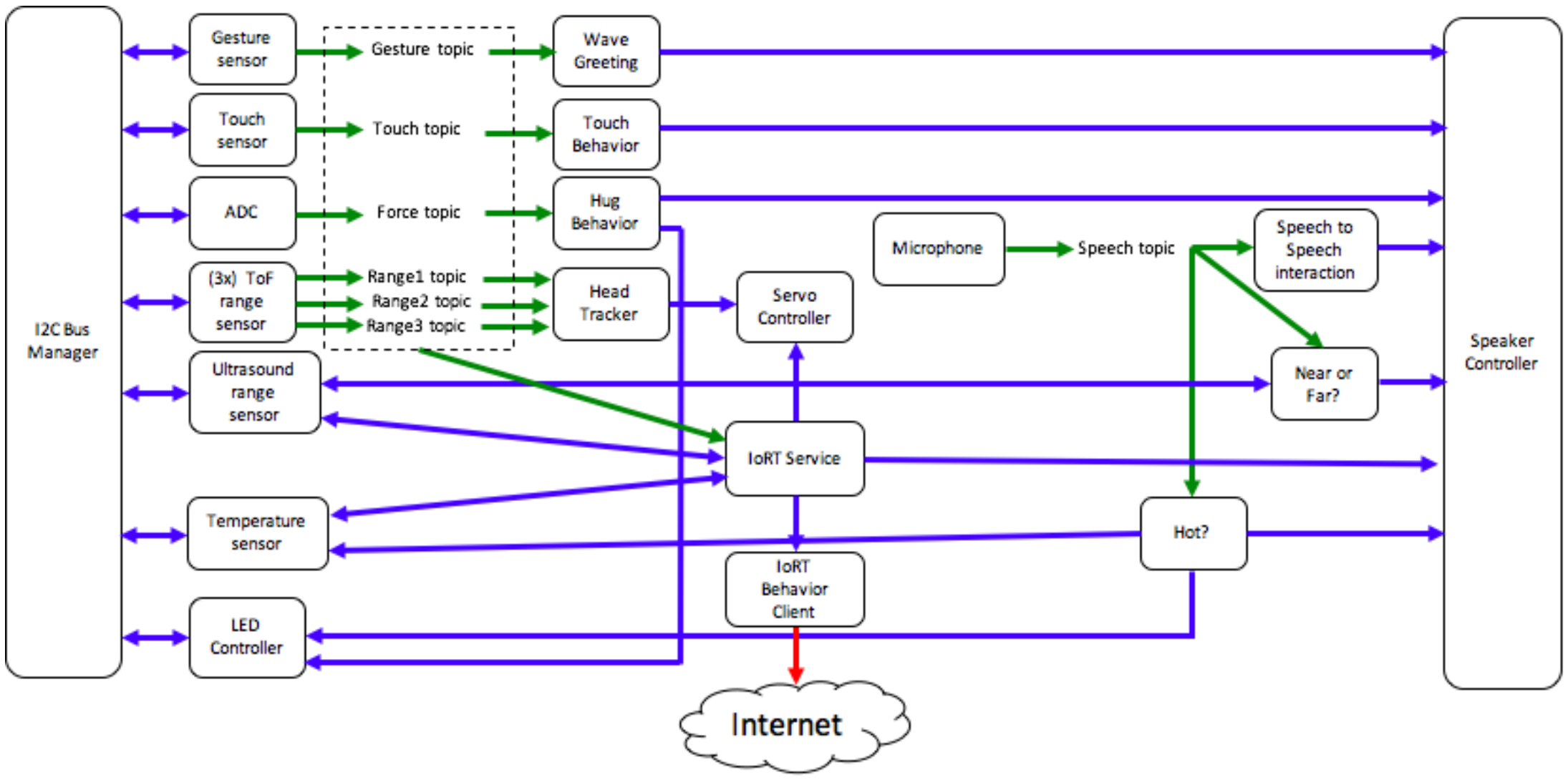

**Fig. 5.** Software diagram of the first prototype as a case study which is run on the GPGPU kit that is mounted on the mannequin.

## 4.2. Software of the our prototype

Fig. 5 shows a block diagram of nodes, messages, and request services between them in this case study. In this prototype, all sensors and controllers use $I^2C$ bus. In this prototype, we used the CMU Pocket Sphinx library [99] in the speech-to-text Node and the Festival library [100] in the text-to-speech Node. There are two ways that our humanoid will trigger a TTS event. The simplest way is to respond to a very specific physical interaction such as patting the humanoid on the back and it says, "Thank you." The second is to respond to the interactor speaking to the humanoid. With a dictionary of words and phrases, the microphone will convert the interactor's speech to strings which are matched to the dictionary and then respond with the corresponding phrase. The method we use for this interaction is a Python daemon that connects to the microphone. The daemon will first measure the ambient volume and once a specified threshold is met, it will begin recording. It records until the silence threshold is met and immediately begins analyzing the converted sound. Once the captured speech is interpreted and matched to the preset text in the dictionary file, the robot responds with an appropriate statement by calling the TTS service.

## 4.3. Internet of Thing of the first prototype

We used Amazon Web Services (AWS) Elastic Compute Cloud (EC2) as the host for our IoT application. AWS gives many easy-to-configure security features at the network level to ensure that our application can be accessed only by necessary users. This combined with Flask's support of SSL-256 and authorization allows our data and access to be secure and safe. Communication over the public Internet introduces an unavoidable increase in latency between physical interaction and commands sent from the remote client application and vice versa. Many uncontrollable factors play a role in causing this latency including, but not limited to ISP inter-AS rules, geography, packet loss, and firewall rules. To mitigate these factors, we hosted our RESTful web service on an extremely high-reliability public network and infrastructure with a 99.99% service level agreement (SLA). When tested we observed a maximum of 3 seconds and a minimum of 1.5 seconds response times. The average response time for IoRT-based interactions is 1.8 seconds.

## 4.4. Hand gesture detection with deep learning on the first prototype

To demonstrate the image understanding subsystem, we use a node that captures images from a webcam and detects the hand gesture (number zero to five). We used OpenCV and the Keras library with the backend of TensorFlow 1.9 to implement a convolutional neural network (CNN). The raw RGB image is captured in 640X360 pixel in 60 fps mode. Then, the raw image is cropped to 128x128 pixels without scaling/rectifying. Next, the cropped image is converted to a grayscale image. Then, it is filtered by a Gaussian filter with a window size of 3x3 and a standard deviation of seven. We used an adaptive thresholding method to convert the filtered image to a binary image (Edge). For finding the threshold for each pixel, we used the weighted average method that the weight comes from Gaussian with the window size of 11x11. Fig. 6 (a) shows examples of the edge images captured from a subject displaying hand sign to the webcam in one set of experiment (Fig.6b). The binary image is inserted into the convolutional neural network. Table 2 shows the detail of the CNN. We used one-hot encoding for 6 states of hands. We used cross-entropy for the cost function. In the training, we used a dropout technique for the fully connected layer. The training accuracy was 97%. We tested the convolutional neural networks as shown in an experiment in Fig. 6b. It detected the hand gesture with an accuracy of 95%. The 2% difference between the training and testing shows that our designed CNN has low variance. The speed was 7 frames per second which is less than our desired speed (10 frames per second). To solve this issue (speed), we could use TensorRT to optimize the network in integer mode. Although TensorRT will increase the frame rate, the accuracy will decrease. Another way to increase the frame rate is to use a lighter CNN; however, the accuracy will decrease. A better way to increase the speed is by replacing NVIDIA Jetson TX2 with a more powerful (higher FLOPS) GPGPU kit. For example, NVIDIA AGX Xavier developer kit is two times faster than NVIDIA Jetson TX2, which increases the frame rate to 14 frames per second (more than desired). In the future works, we will use a combination of better GPGPU and TensorRT, i.e., we will explore a deeper CNN in integer mode, using TensorRT for NVIDIA AGX Xavier.

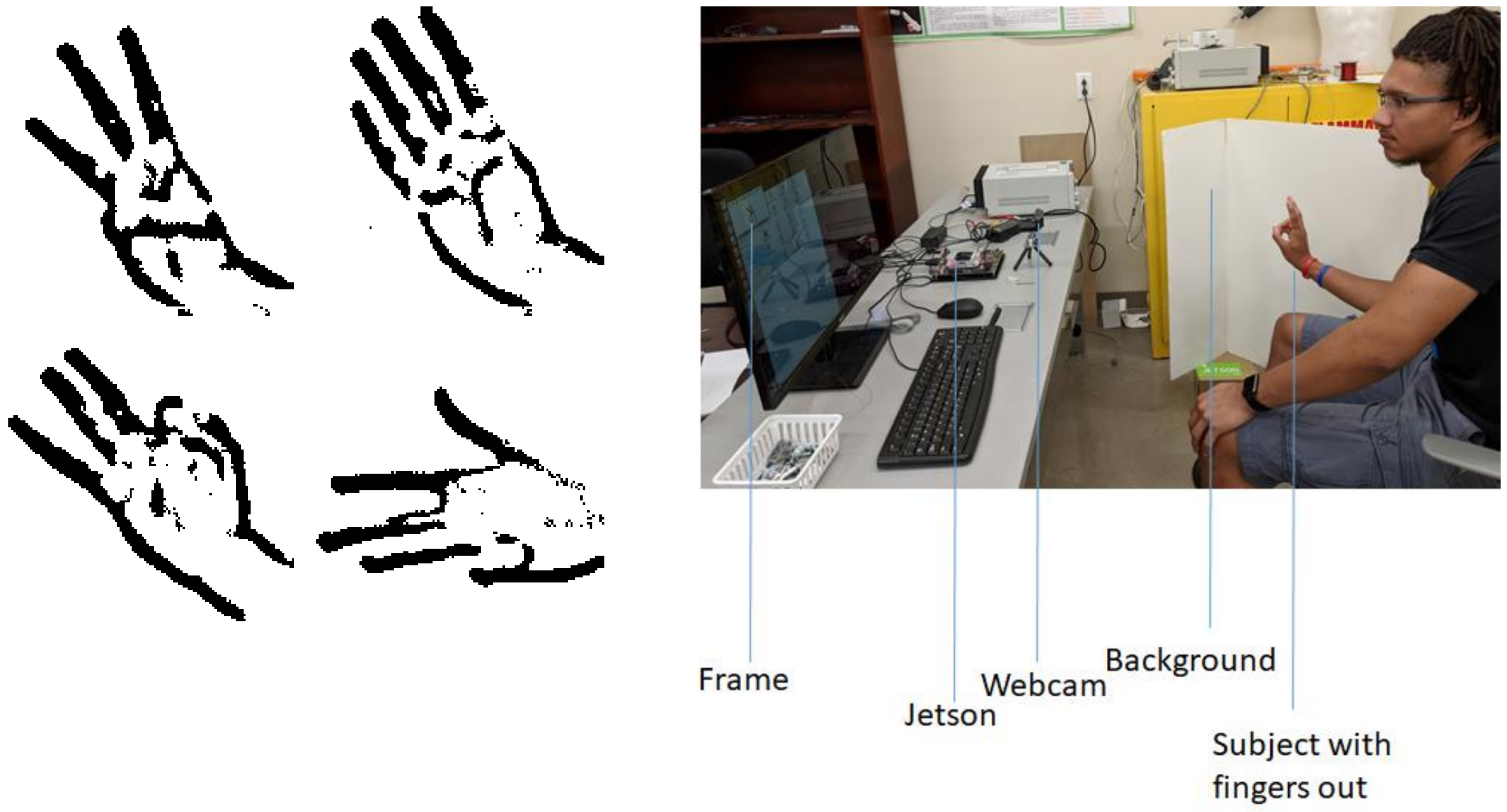

**Fig.6.** Hand gesture recognition using deep learning and NVDIA Jetson TX2: (a) Examples of edge images, (b) experimental setup.

Table 2. Layers of the convolution neural network (CNN) for detecting hand gesture with deep learning.

| Number | Type | Filter size | Number of Filter | Activation | Input Size | Output Size |
|---|---|---|---|---|---|---|
| 1 | Convolution | 3 x 3 | 8 | ReLU | 128 x 128 x 1 | 128 x 128 x 8 |
| 2 | Pooling | 2 x 2 | 1 | Max | 128 x 128 x 8 | 64 x 64 x 8 |
| 3 | Convolution | 3 x 3 | 16 | ReLU | 64 x 64 x 8 | 64 x 64 x 16 |
| 4 | Pooling | 2 x 2 | 1 | Max | 64 x 64 x 16 | 32 x 32 x 16 |
| 5 | Convolution | 3 x 3 | 32 | ReLU | 32 x 32 x 16 | 32 x 32 x 32 |
| 6 | Pooling | 2 x 2 | 1 | Max | 32 x 32 x 32 | 16 x 16 x 32 |
| 7 | Convolution | 3 x 3 | 64 | ReLU | 16 x 16 x 32 | 16 x 16 x 64 |
| 8 | Pooling | 2 x 2 | 1 | Max | 16 x 16 x 64 | 8 x 8 x 64 |
| 9 | Convolution | 3 x 3 | 128 | ReLU | 8 x 8 x 64 | 8 x 8 x 128 |
| 10 | Pooling | 2 x 2 | 1 | Max | 8 x 8 x 128 | 4 x 4 x 128 |
| 11 | Convolution | 3 x 3 | 256 | ReLU | 4 x 4 x 128 | 4 x 4 x 256 |
| 12 | Pooling | 2 x 2 | 1 | Max | 4 x 4 x 256 | 2 x 2 x 256 |
| 13 | Convolution | 3 x 3 | 512 | ReLU | 2 x 2 x 256 | 2 x 2 x 512 |
| 14 | Pooling | 2 x 2 | 1 | Max | 2 x 2 x 512 | 1 x 1 x 512 |
| 15 | Fully Connected | 512 x 128 | 1 | ReLU | 1 x 1 x 512 | 1 x 1 x 128 |
| 16 | Fully Connected | 128 x 6 | 1 | Softmax | 1 x 1 x 128 | 1 x 1 x 6 |

## 4.5. Time-based Sensing Module Interface

The time-based sensing module means the sensors read input data periodically. The time-based sensing module includes modules of gesture sensing, touch sensing, force sensing, and range sensing.

### 4.5.1. Touch Sensing Module

We use an Adafruit MPR121 capacitive touch sensor, which communicates via $I^2C$ bus. The touch publisher node requests the $I^2C$ manager node to read data from the touch sensor periodically. When the touch sensor node receives data from $I^2C$ manager node, it publishes the data to the touch topic. If a touch has occurred on the touch sensor, the touch node will send a pre-defined phrase such as ‘thank you’ to the text to speech service node. The conceptual representation of the touch sensing module is shown in Fig. 7. Inside the dash lines is the process of getting touch data. To get the data from the touch sensor, user can send a remote command “read touch” using the user interface client node to the cloud. The IoRT server node subscribes to the touch topic. When the IoRT behavior node detects the command, it will request the IoRT server node to get the touch data from the touch topic. The codes for each segment are shown in Fig. 7 (b) to (e).

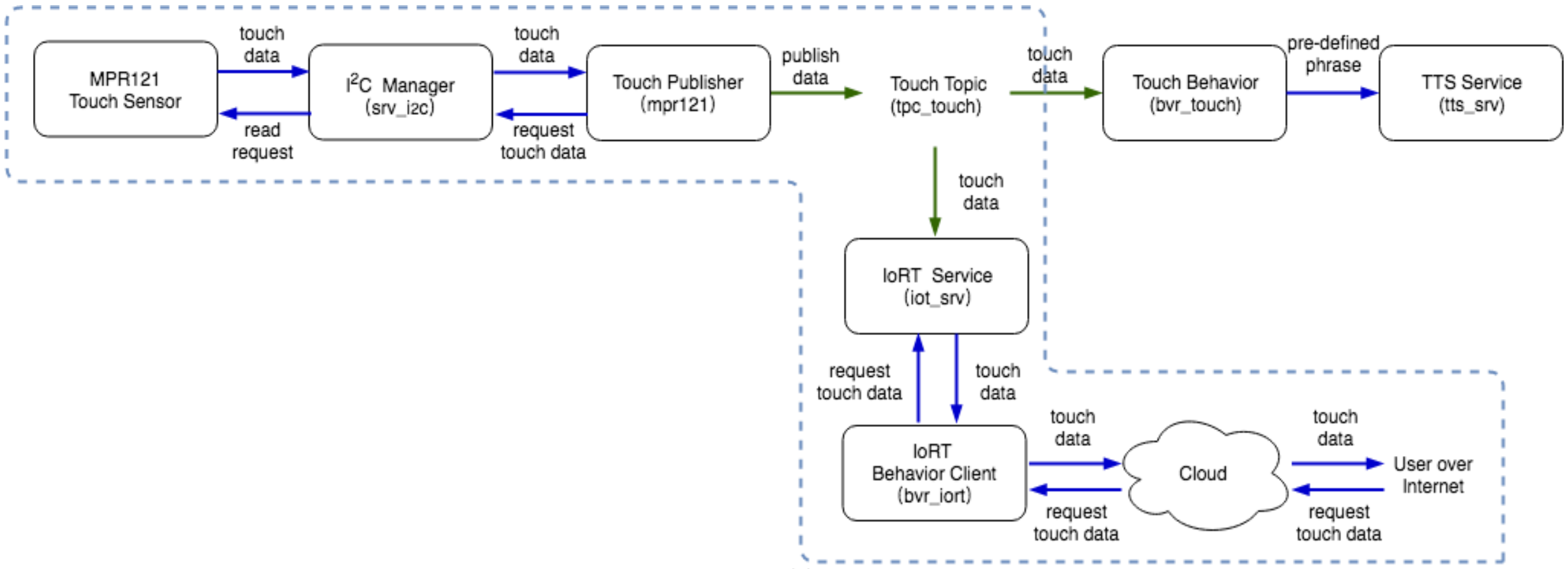

(a)

```
mpr121_publisher.cpp:
#include "ros/ros.h"
#include "std_msgs/UInt8.h"
int main(int argc, char **argv)
{
   ros::init(argc, argv, "mpr121_publisher");
   ros::NodeHandle n;
   ros::Publisher touch_pub = n.advertise<std_msgs::UInt8>("tpc_touch", 1);
   ros::Rate loop_rate(1);
   ROS_INFO("MPR121 initialized successfully.");
   while (ros::ok()){
      std_msgs::UInt8 msg;
      ...
      touch_pub.publish(msg);
      ros::spinOnce();
      loop_rate.sleep();
   }
   return 0;
}
```

(b)

```
hbs_api,py:
app = Flask(__name__)
@app.route('/api/getcommand')
def get_command()
@app.route('/api/getdata')
def get_data()
@app.route('/api/setdata')
def set_data()
@app.route('/api/readtouch')
def readtouch():
   logger.info("Request to read touch sensor has been made.")
   with open(COMMAND_FILE_PATH, 'wb') as cmd_file:
      logger.info('Writting 0x06 to command file.')
      cmd_file.write(b'\x06')
   return b'\xFF'
```

(c)

```
iot_srv.cpp:
uint8_t global_memory_for_touch = 0;
void touchCallback(const std_msgs::UInt8 msg)
{
   global_memory_for_touch = msg.data;
}
switch (req.command) {
   case 6: {
      res.data = global_memory_for_touch;
      return true;
   }
      default:
         break;
   }
int main(int argc, char **argv) {
   ros::init(argc, argv, "iot_srv");
   n = ros::NodeHandlePtr(new ros::NodeHandle);
   ros::Subscriber touch_sub = n->subscribe("tpc_touch", 1, touchCallback);
   ROS_INFO("ROS IoT Service has started.");
   ros::spin();
   return 0;
}
```

(d)

```
bvr_iort.cpp:
Begin sending HTTP GET to Request endpoint;
void begin(ros::NodeHandle &n) {
   ros::ServiceClient iot_client = n.serviceClient<hbs2::iot>("iot_srv");
   hbs2::iot srv_iot;
   rest_req(resp, get_cmd_uri);
   switch (m_pBuffer[0]) {
      case 6: {
         send request to iot_srv
         req.request.command = 6;
         check for success and return data back to api client;
         create rest_req(resp, set_resp_uri);
      }
   }
}
int main(int argc, char *argv[]) {
   ros::init(argc, argv, "iort");
   ros::NodeHandle n;
   ros::Rate loop_rate(1);
   while(ros::ok()) {
      begin(n);
      ros::spinOnce();
   loop_rate.sleep();
   }
   return 0;
}
```

(e)

**Fig.7.** Conceptual representation of touch sensing module: (a) inside dashed lines is the process of getting raw touch data, (b) code of touch publisher node, (c) code of the API in the cloud, (d) code of IoRT service node, and (e) code of IoRT behavior client node.

#### 4.5.2. Gesture Sensing Module

In this module, the gesture publisher node requests the $I^2C$ manager node periodically to read data from the gesture sensor then publishes the data to the gesture topic. The wave greeting node, which subscribes to the gesture topic gets the data from the gesture topic. When a wave has occurred, the wave greeting node sends a pre-defined phrase such as 'Hello' to the text to speech service node. We can also get gesture data by using IoRT communication. Fig. 8 (a) shows the conceptual representation of gesture sensing.

#### 4.5.3. Force Sensing Module

We use two force resistive sensors and an analog-to-digital converter (ADC) in this module. The force publisher node requests the $I^2C$ manager node periodically to read data of all channels in ADS1115 hardware. Then the force publisher node publishes the data to the force topic. The hug node subscribes to the force topic to read the force data. If both the right and left force sensors data reach a certain threshold at the same time, then the hug node requests the LED driver node to change the LED color and send a pre-defined text to the text to speech service node. This indicates a hugging action. The conceptual representation of the force sensing module is presented in Fig. 8 (b).

#### 4.5.4. Range Sensing Module

To track the position of the nearest object, three ToF range sensors are used. The range publisher node requests the $I^2C$ manager node to read data from each VL53L0X sensor and then publishes the data to the track topic. We use a tracker node to subscribe to the tracking topic.. The tracker node finds the minimum distance measured from each sensor. If minimum distance is not the center distance, the tracker will request the servo service node to move toward the sensor with the minimum measurement. Fig. 8 (c) is the conceptual representation of the range sensing module.

### 4.6. Event-based Sensing Module

The event-based sensing modules mean the sensors nodes read input data after receiving a request from another nodes. The event-based sensing module includes ultrasound range sensing module and temperature module.

#### 4.6.1. Ultrasound Range Sensing Module

The SRF02 sonar proximity sensor communicates via $I^2C$ bus. The sonar sensor node is a service node that reports the range when a call to its advertised service is made from the Internet of Robotic Things (IoRT) node. The sonar service node requests the $I^2C$ manager node to write a "real ranging mode-result in centimeters" command to the SRF02 device and waits the allotted time for the measurement to be made. Then the sonar service node reports the resulting range in centimeters to the IoRT service node. The conceptual representation of the ultrasound range sensing module is shown as Fig. 9 (a).

#### 4.6.2. Temperature Module

We use the MCP9808 sensor to measure the temperature of the environment around the humanoid robot. The sensor, which communicates via $I^2C$ bus, converts temperatures to a digital word data (16 bits). This data is then manipulated to represent the temperature in degrees Fahrenheit. The temperature service node reports the temperature when a call to its advertised service is made by IoRT service node. Fig. 9 (b) shows the conceptual representation of the temperature module.

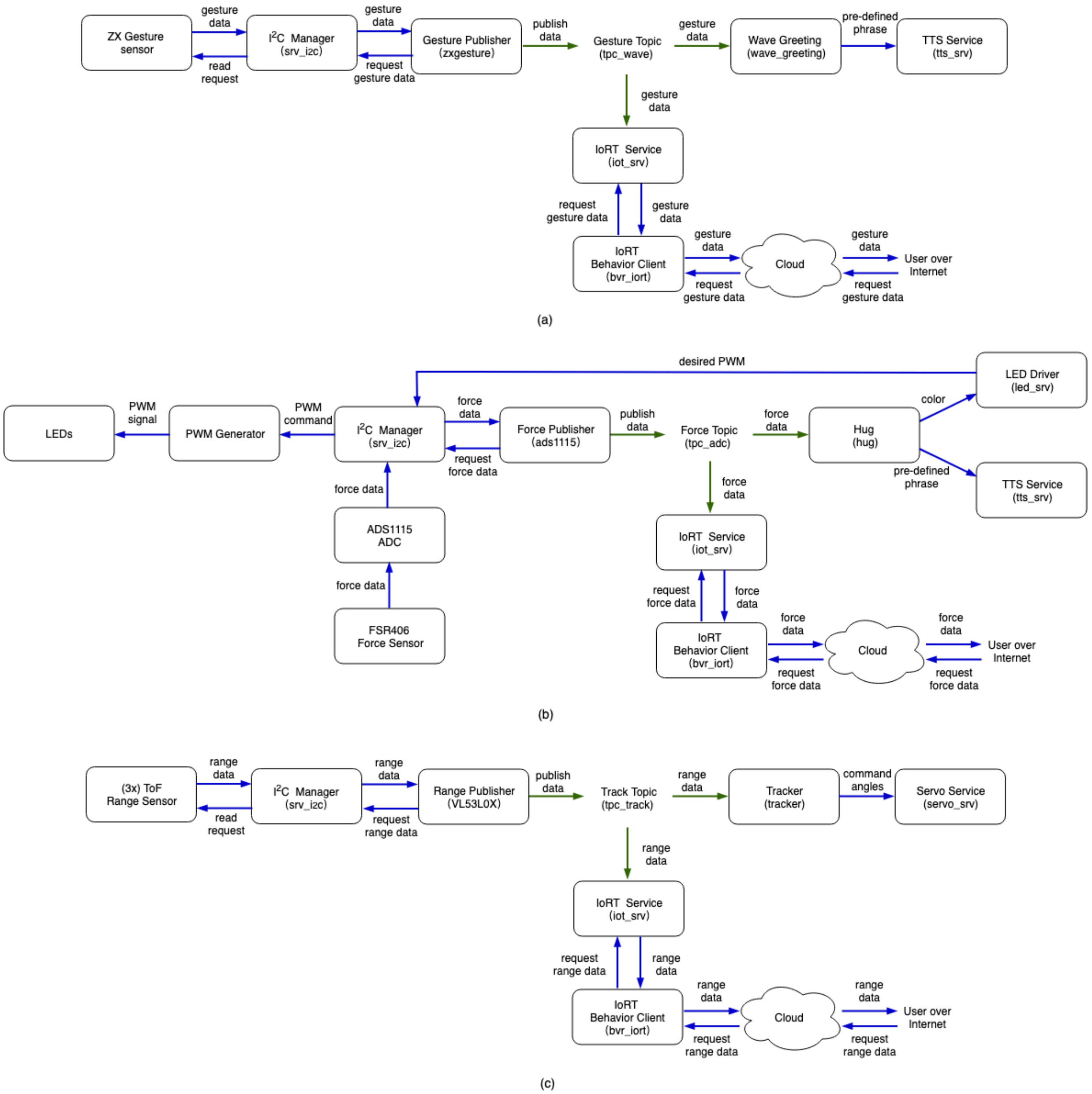

**Fig. 8.** Conceptual Representation of: (a) Gesture Sensing Module, (b) Force Sensing Module, and (c) Range Sensing Module.

## 4.7. Speech to Speech Interaction Module

When an audio is captured by the speech publisher node, the speech to text (STT) node will interpret the raw audio data to text using CMU Pocket Sphinx and publish the text to the speech topic. The speech to speech interaction node subscribes to the speech topic with the frequency of one second. Once the speech to speech interaction node reads a new message from the speech topic, it will check whether the text of the message is in the dialog database, which is defined in the config.txt. If the text exists in the database, then the speech to speech interaction node will send the pre-defined response in the database to the text to speech node. The conceptual representation of the speech to speech interaction module is shown as Fig. 9 (c).

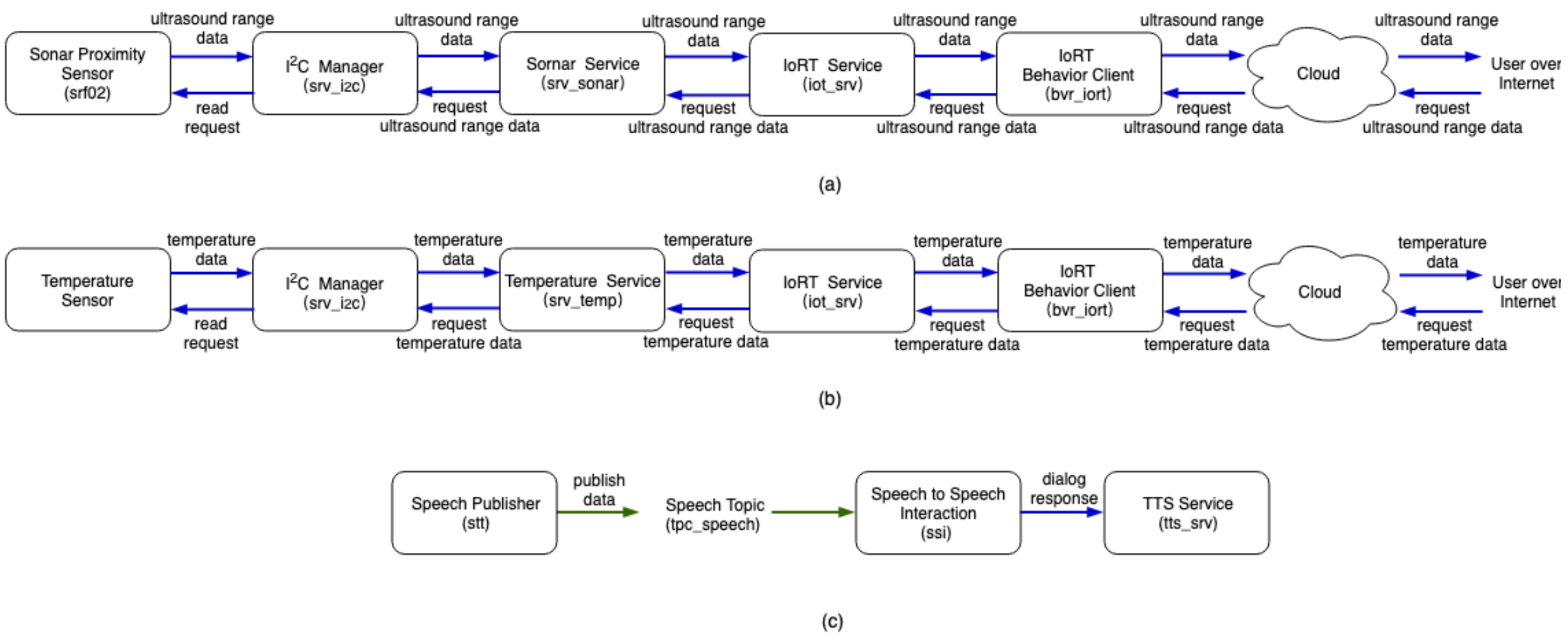

**Fig. 9.** Event based sensing module- Conceptual representation of: (a) Ultrasound Range Sensing Module, (b) Temperature Module, (c) Speech to Speech Interaction Module.

Comparison of our hardware and software features are provided in supplementary files (Table S3, S4 and S5) by considering different features and to provide a more detailed explanation.

## 4.8. Experimental result of the prototype

Fig. 10 illustrates the experiments on the first prototype, the illustrated wearable vest. In Fig. 10 (a), a human talks with the robot for testing the voice interactions. The gesture sensor (Fig. 10 (b) shows the test on the gesture) requires a close distance to detect the hand motion. Fig. 10 (c) shows testing the ultrasonic range sensor using a paper along the way. Fig. 10 (d) shows testing the tracking behavior node by moving a hand over the head area. The testing touch sensor is demonstrated in Fig.10 (e) by tapping the back area (copper pads). The hug behavior testing illustrated in Fig. 10 (f), where a user presses the side sensors in the vest. These sensors are located on the left and the right side of the vest. Video illustration of these tests is available in the HBS lab YouTube channel in the following link: https://youtu.be/8RuS-vbxqPk. The objective in this case is to show the functionality of the first low-cost prototype that was used to test the proposed software architecture. Other measurements are also performed on the current prototype. Fig. 11 (a) shows one set of measuring distance with the ultrasonic range sensor, which was done by placing a paper pad at a different position from the sensor. Fig. 11 (b) shows one set of distance measurements with the ToF range sensor. It can be seen that the ultrasonic sensor detects a far distance from 50 to 250 cm, whereas the ToF sensor responds to a shorter distance from 5 to 20 cm. Fig. 11 (c) shows the sensitivity of the FSR sensor, which indicates that as the applied load is increased, the voltage across the FSR is decreased. This test was done by applying a calibrated weight on the FRS sensor and measuring the output voltage. It was discussed at the beginning that the FSR resistance decreases its resistance as the load increases. Fig. 11 (d) shows the time response of the temperature sensor when it contacts with a warm/cold object. The red line indicates warm object ($40^0$C) and the blue color is the cold object ($22^0$C). In both cases, as time goes by, the temperature decays and reach to steady state within 500-600 seconds.

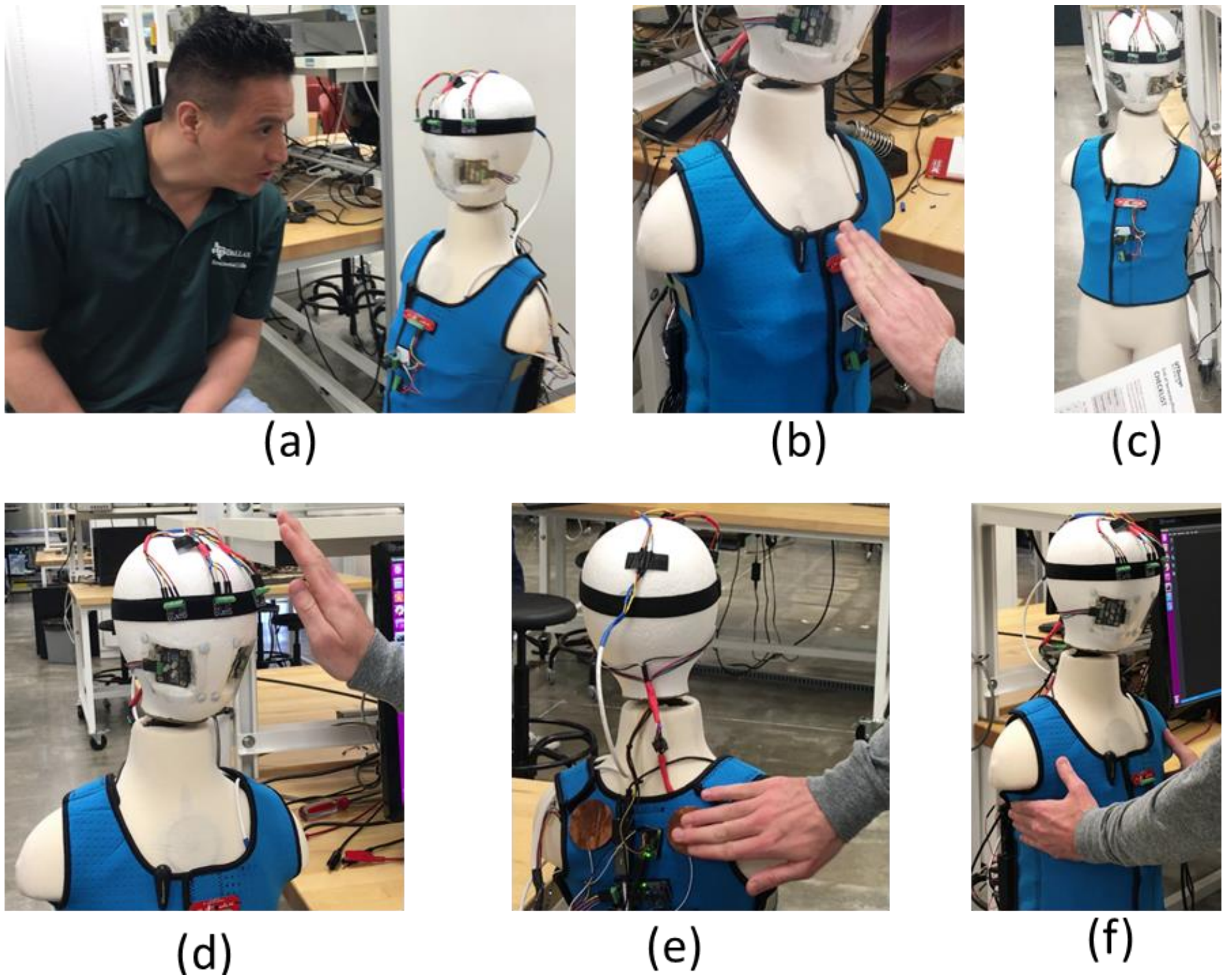

Fig. 10. Example of testing the modular wearable vest: (a) speech interaction, (b) gesture recognition, (c) ultra sound range sensor, (d) time of flight range sensor, (e) capacitive touch sensor, and (f) force sensitive resistor.

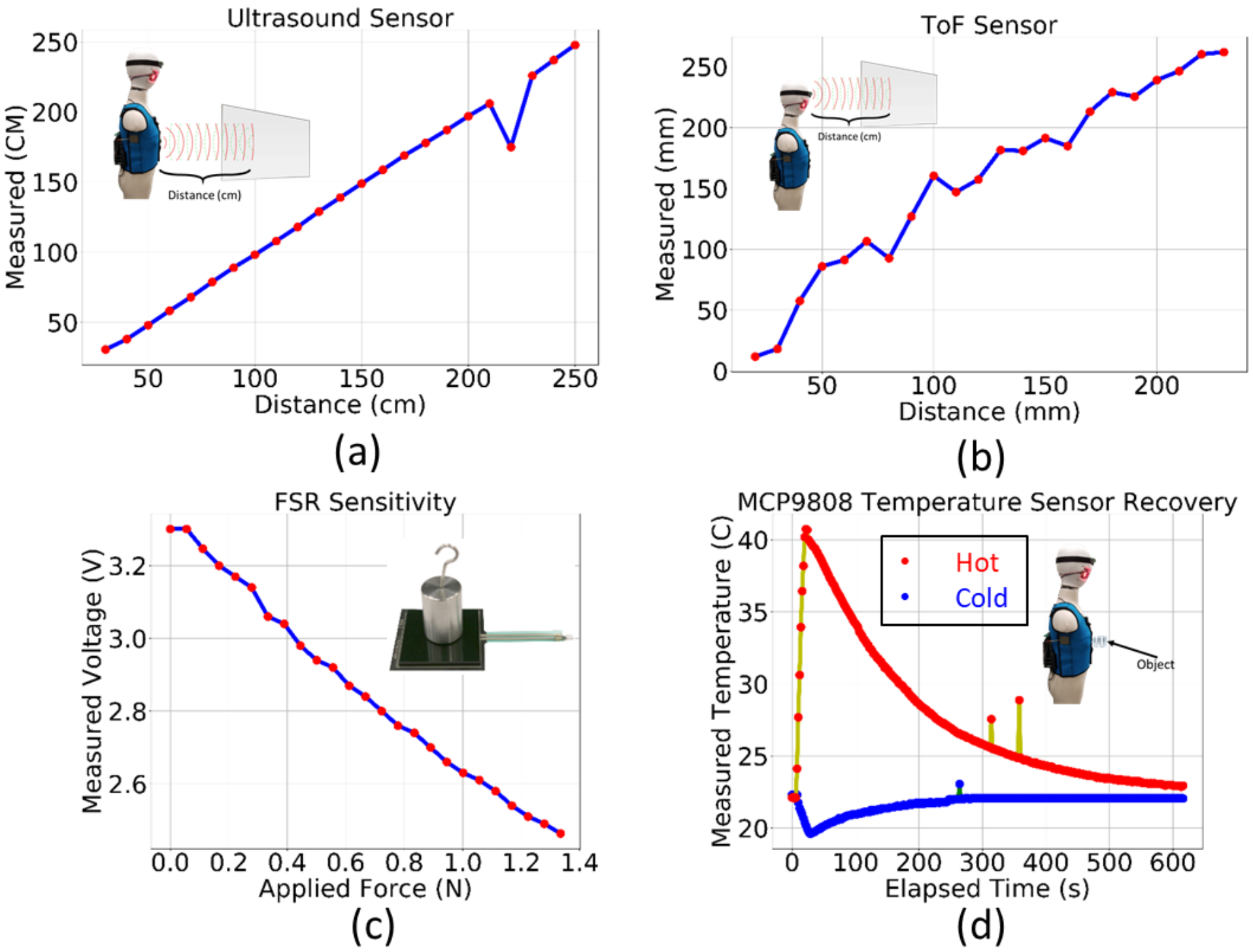

Fig. 11. Experimental results of testing sensors: (a) ultrasound range sensor, (b) ToF range sensor, (c) FSR sensor and (d) temperature sensor.

Interaction through speech is one of the most important media for social humanoid robots. To test the ability of vocal interaction, a person talks to the robot (Fig. 10 (a)) as a preliminary test, and the robot recognizes the voice through its TTS and the software discussed earlier. The interaction consists of three parts. The first is the speech to the text. We performed experiments with 10 subjects to test the speech. Subjects were told to repeat certain phrases 5 times. Fig. 12 shows the accuracy of phrases when 10 subjects repeated each sentence/phrase 5 times (total of 50 times). We followed IRB protocol when conducting experiments with human subjects. CMU Pocket Sphinx failed to detect the word "time" and successfully detected all of "what is your" phrases. This experiment shows that we should use a new technique to overcome this issue. We used the current hardware /software for the speech because it did not require Internet connection and they are available for free use. We plan to use an end-to-end neural network (deep learning) in future work. The accuracy of online services like IBM is better; however, it may be slower because of Internet communication, which should be avoided in social robots. The second part is mapping the recognized text to the corresponding response text in the database to create dialogues such as:

- Dialogue 1:
  - Human: "What is your name?"
  - Robot: "My name is H B S 2"
- Dialogue 2:
  - Human: "What is your favorite color?"
  - Robot: "My favorite color is blue. My vest is blue."
- Dialogue 3:
  - Human: "What is the temperature?"
  - The robot collects temperature sensor data and passes it to the TTS Service Node
  - Robot: "The temperature is 72 degrees"

The third part is converting the response text to the speech. The resulting speech is clear; however, it is artificial because the software combines the phones to play. The more natural way is to have a database of sentences; however, it will be limited as it requires a high volume of permanent memory (SSD storage). Also, it adds more delay to search for the response files. Cloud computing [46, 49, 101] is another method to consider, however, it causes the robot to stop working when the Internet connection fails that is not desired for social robots.

Responding/reaction times are very important in social humanoid robots. If a robot shows long delays, the humans that interact with the robot feel awkward and they lose eagerness to talk with the robot. Fig. 13 shows some examples of the delay time in experiments. Reaction times to touching (capacitive) and hugging (FSR) are negligible. Shaking head by requesting over the Internet takes less than half a second. This demonstrates the success of our goal. Reading temperature takes about half a second. These delays come from the fact that $I^2C$ bus share with many sensors. Regardless, this much delay is acceptable. Reading the ultrasound sensor and requesting the text to speech via Internet takes about 1.5 s and 2.5 s respectively. These results are not acceptable and should be fixed in future works.

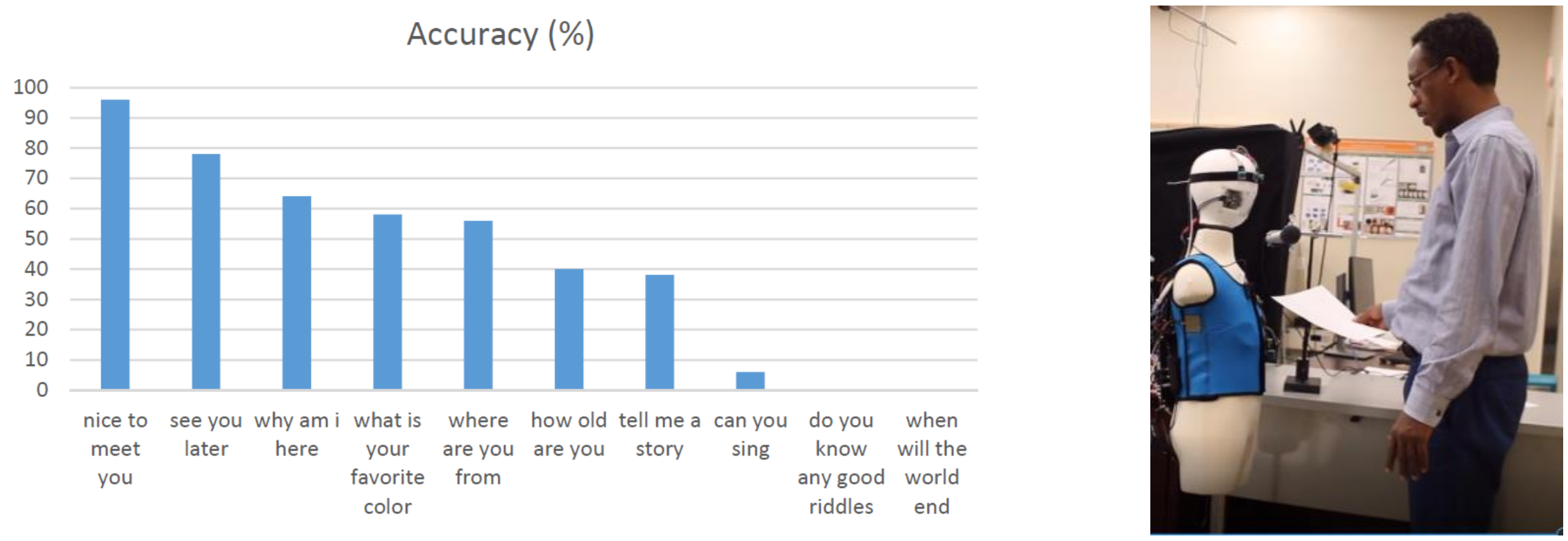

**Fig.12.** Accuracy of speech to text when each phrase repeated 50 times (Left) and the experimental setup (right). Subjects are asked to stand next to the robot with sensor vest and read the phrases 5 times and the recorded voice is analyzed.

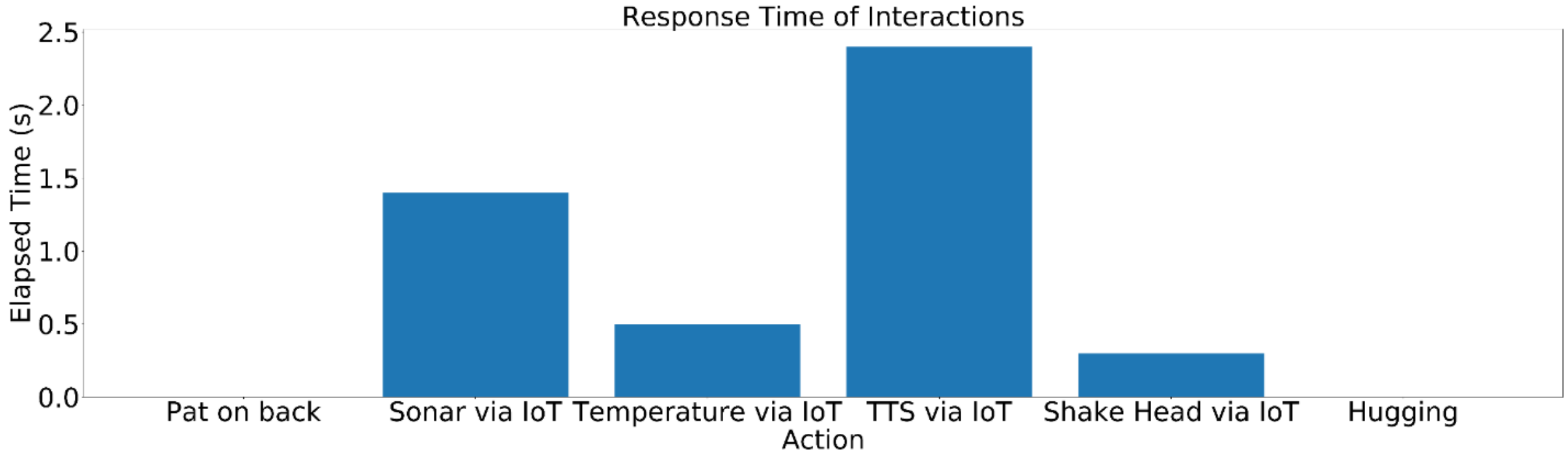

**Fig.13.** System response time during interactions

# 8. Discussion

To add a new sensor, we need to decide which bus we should to use for communication between the sensor and GPGPU, for example, the I2C, SPI, or UART. If we decide to use $I^2C$ to communicate with the sensor, we must connect the sensor to an $I^2C$ bus and create either a sensor server or a publisher node in ROS. If the sensor is unable to communicate via $I^2C$, we need to create a bus manager of the type, which the sensor is compatible. When a new behavior node is created, it can be subscribed to one or more topics.

As mentioned in the introduction, social robots should respond in real-time and in a natural way to not create an awkward or unnatural feeling when they interact with humans. The first prototype can be improved greatly by replacing the speech to text and text to speech subsystem with other options. The latency of Internet communication can be addressed by utilizing fog computing and concurrency. The latency from the deep learning will be solved by using a more powerful developing kit.

We think that the social robot should do most of the computation locally. For example, we could record the voice and send it to a cloud robotic and receive the text back. We chose to use local speech to text even though we know that the accuracy is much lower than the cloud services. With local computing, we are able to eliminate the latency that comes from Internet communication. Most importantly, if the social robot depends entirely on the Internet, it will stop functioning when the Internet connection is lost. For communication with people that physically do not have access to the robot, we designed the IoT node; the node connected to a cloud server. To decrease the latency and increase safety, we can move to edge computing servers; however, that subject needs further study. Currently, we used very few sensors, but when increasing the number of sensors, we could consider the distributed software architecture and keep the current architecture at the top of the network (hierarchy).

Noise and disturbance are unavoidable in electronic interfaces. These are some of the challenges in wearable vests for humans because human signals are analog and low power such as EMG, ECG, ENG, etc. However, noise and disturbance in digital buses are negligible especially in the case of USB, $I^2C$, and SPI bus. Therefore, we only used digital sensors and actuators in our hardware architecture. The digital sensors refer to the combination of the sensitive elements and the signal processing unit that comes with the sensors or the analog to digital convertors. For example, the FSR and the ADS1115 combined is the digital force sensors used in the vest.

Artificial skins for humanoids are proposed in prior works [102, 103]. We have also tried piezoelectric sensors embedded in flexible skin for social robots [91, 104, 105] . However, casting the materials, embedding sensors permanently in elastomers and integration require more involved processes. Still such flexible artificial skins can be used in the vest, as the sensing parts are one aspect of the current work. In contrast to artificial skin, the wearable vest presented in this paper is more convenient to implement low-cost sensors (as the fabric can be stitched, weaved and adjusted) and is able to allow developers to add new sensors on it or remove unused sensors from it. Another advantage of our sensor vest over artificial skin is that the vest can be repeatedly used on different robots since it is wearable and fits many exist social humanoid robots. The parts used in the prototype of the vest are changeable and work independently. Currently, we are using some low-cost hardware components for our vest to make it affordable and they can be replaced based on the consideration and balancing the requirement of kinematic performance and the cost.

This paper presents a hardware architecture as well as a software architecture for wearable vests for social humanoid robots. The architecture is modular and independent of shapes, materials, and components. We showed a case study for this architecture and its performance. We used low-cost sensors, actuators, vest, and GPGPU kit. All elements can be replaced by a better one without changing the architecture. For example, the vest in the first prototype may avoid kinematic abilities of some robots; however, a tight fit and a muscle shirt (Fig.14a-e) that can be cut in some portion can be employed. This can be one design variable of the vest, such that what portion of the robot should be covered or not. The other issue is overheating if the robot is entirely covered by the sensor vest. This issue can be easily addressed by considering meshed fabrics (Fig.14 c) that are highly elastic for a tight fit or porous fabric that can be employed for better heat flow. This can be a second design variable of the vest that designers select. In future works, we will use better components (as discussed in the table S5) and replace the vest of the first prototype with a lightweight porous, thin, and stretchable fabric. Also, for some robots, we can cut the fabric (like Fig,14d) in a way that will not occlude the robot's joints movement and cooling fans can be added to prevent overheating. In some applications related to children or social interaction with ASD children, the entire wiring and controller should be covered for safe human-robot interaction (HRI). In order to do this, a second layer should be used like the one shown in Fig.14 (e). Therefore, a multilayer sensor vest can be considered to address the safety issues of such robotic vest design. In the case of 3D printed robots, one idea to overcome the issues of wiring is to 3D print the wires together with the structural parts as presented in[106]. But this will be beyond the scope of this paper.

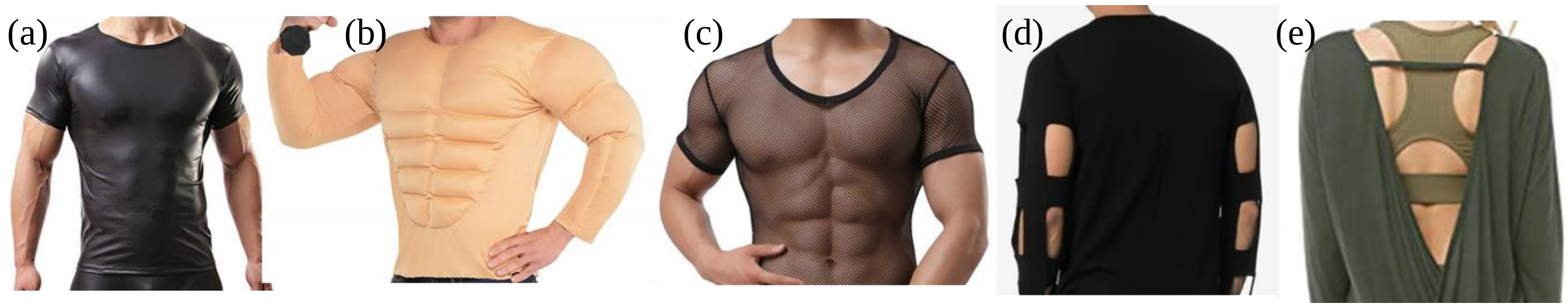

Fig.14 Garment design for improvement of the vest: (a) tight fit with elastic fabric, (b) full body covering tightly, (c) meshed or fishnet shirt design, (d) cut out in some areas and (e) multiple layers for safety protection. (Original Figures a, b, and e are from amazon.com; c is from wish.com; d is from brownfashion.com).

Here, we show the vest with standard robotics actions with cameras and sensors, such as hand gesture recognition or speech recognition. The goal of this research is designing a general-purpose hardware and software architecture to be used as a starting point of designing complex behaviors. For example, it helps psychologists to design therapy sessions for children with autism spectrum disorders (ASD). In this example, psychologists do not need to know about low-level nodes and hardware.

Safety, security, and privacy are common concerns in all IoT based systems. To address safety concerns, the architecture requires adding some administrative and/or orchestration nodes to control all other nodes to react to node failures that occur. The security and privacy of the robots directly depend on the cloud that runs the host. Therefore, they need a completely different study.

Medical care and health care represent one of the most attractive application areas for social robots [107-109]. The modular software architecture of the IoT vest makes it easy to develop more complex functions based on the basic sensing modules presented in our work without changing the nodes and the architecture of each module. The designed architecture has the potential to give rise to many medical applications such as in the treatment of attention in autistic children [110-112], interacting with babies and toddlers [113], remote health monitoring [114], fitness programs, and elderly care [115, 116]. Compliance with treatment and medication at home and by healthcare providers is another important potential application. Tourism [117, 118] and education [119-121] are other areas that can benefit from this software architecture.

# 7. Conclusions

In this paper, we showed an IoT-based wearable sensor vest with GPGPU for social humanoid robots, highlighting the hardware and software architectures. We showed the vest in a robot that is connected to the wireless Internet to send or receive information and commands from a virtual private server (VPS) or cloud. We designed an open-source modular software architecture for social humanoid robots. The proposed software architecture can be used for any social robots that have a GPGPU, ROS, and SPI/$I^2C$ buses. The software architecture consists of low-level nodes, sensor nodes, and behavior nodes. The software architecture manages the communication buses. The architecture provides the IoT node to communicate with other people and robots via the Internet. The API is also capable of accepting and differentiating between commands to multiple robots. Developers can easily add/remove/update nodes because it is designed to be modular as much as possible. The proposed software architecture provides provisions for audio-visual interaction, which consists of speech-to-text, text-to-speech, and image understanding. This architecture helps developers to write complex behaviors for social interaction by isolating the behavior nodes from other nodes.

In developing the software architecture for social humanoid robots, we have encountered many challenges. Many sensors are connected to a bus and the bus communicates with each sensor in a time-based manner. Increasing the number of sensors is challenging due to management issues. We will consider distributed and hierarchy-based communication systems for future works. Improving the speech-to-text and text-to-speech nodes are the most challenging works, which should be priorities for future work to improve vocal interaction. Latency results from communication over the public Internet is another challenge that separate studies should address.

**Acknowledgment**: We would like to thank the senior design team: Sharon Choi, Manpreet Dhot, Mark Cordova, Luis Hall-Valdez, and Stephen Brooks for design and development of the initial design.